\documentclass[conference]{IEEEtran}
\IEEEoverridecommandlockouts
\usepackage{cite}
\usepackage{amsmath,amssymb,amsfonts}
\usepackage{algorithmic}
\usepackage{graphicx}
\usepackage{caption}
\usepackage{subcaption}
\usepackage{textcomp}
\usepackage{xcolor}
\usepackage{placeins}

\usepackage{cuted}
\usepackage{capt-of}
\usepackage{color}
\usepackage{enumitem}
\usepackage{wrapfig}
\usepackage{flushend}
\usepackage{url}

\def\BibTeX{{\rm B\kern-.05em{\sc i\kern-.025em b}\kern-.08em
    T\kern-.1667em\lower.7ex\hbox{E}\kern-.125emX}}

\makeatletter
\newcommand{\linebreakand}{%
\end{@IEEEauthorhalign}
\hfill\mbox{}\par
\mbox{}\hfill\begin{@IEEEauthorhalign}
}
\makeatother

\begin{document}
\setlength{\parindent}{0pt}

\title{Segformer++: Efficient Token-Merging Strategies for High-Resolution Semantic Segmentation}

\author{
\IEEEauthorblockN{Daniel Kienzle}
\IEEEauthorblockA{
\textit{University of Augsburg}\\
Augsburg, Germany \\
daniel.kienzle@uni-a.de}
\hfill
\and
\IEEEauthorblockN{Marco Kantonis}
\IEEEauthorblockA{
\textit{University of Augsburg}\\
Augsburg, Germany \\
marco.kantonis@uni-a.de }

\and

\IEEEauthorblockN{Robin Schön}
\IEEEauthorblockA{
\textit{University of Augsburg}\\
Augsburg, Germany \\
robin.schoen@uni-a.de }
\hfill
\and
\IEEEauthorblockN{Rainer Lienhart}
\IEEEauthorblockA{
\textit{University of Augsburg}\\
Augsburg, Germany \\
rainer.lienhart@uni-a.de}

}

\maketitle

\begin{strip}
    \vspace{-1.8cm}
    \centering
    \includegraphics[width=\textwidth]{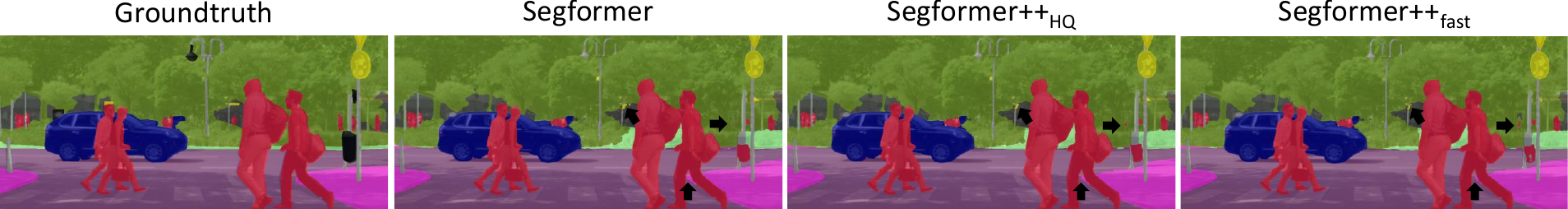}
    \captionof{figure}{Qualitative comparison of our efficient models. Even though our models are significantly faster, nearly no quality degradation can be observed. Small differences are highlighted with black arrows.}
\label{fig:feature-graphic}
\vspace{-0.41cm}
\end{strip}

\begin{abstract}
Utilizing transformer architectures for semantic segmentation of high-resolution images is hindered by the attention's quadratic computational complexity in the number of tokens. 
A solution to this challenge involves decreasing the number of tokens through token merging, which has exhibited remarkable enhancements in inference speed, training efficiency, and memory utilization for image classification tasks. 
In this paper, we explore various token merging strategies within the framework of the Segformer architecture and perform experiments on multiple semantic segmentation and human pose estimation datasets. 
Notably, without model re-training, we, for example, achieve an inference acceleration of 61\% on the Cityscapes dataset while maintaining the mIoU performance. 
Consequently, this paper facilitates the deployment of transformer-based architectures on resource-constrained devices and in real-time applications.
\end{abstract}


\vspace{-0.225cm}
\section{Introduction}
Nowadays, real-life images, e.g. captured with mobile phone cameras, are of high resolution.
However, most computer vision applications still work with low resolution images due to the high computational cost. 
Especially vision transformers (ViT) \cite{visiontransformer} struggle with the quadratic complexity of the self-attention mechanism in high resolutions images. 
Consequently, most ViT based architectures perform a significant downsampling of the input image resolution. 
While this might not be a huge problem for tasks like image classification, it is a major drawback for dense pixel tasks like semantic segmentation, monocular depth estimation or human pose estimation. 
As a result, modern architectures like Segformer \cite{segformer} have been developed to handle higher resolution images. 
While Segformer is based on the vision transformer architecture, it introduces an efficient attention mechanism to reduce the computational burden of the attention mechanism.  
Thus, it is a relatively efficient but still powerful architecture for dense pixel tasks that, in contrast to purely convolutional architectures, still makes use of the global receptive field of the attention mechanism. 
However, for real-life applications, even more efficient architectures are needed to deal with high resolutions and to enable real-time computations on edge-devices. 
Thus, we adjust the recently proposed token merging \cite{tokenmerging} strategy, introduced for increasing the efficiency of vision transformers in image classification, to the Segformer architecture. 
Even though it is not straightforward to apply token merging out-of-the-box due to the frequent use of convolutions in Segformer, we adjust the algorithm for this architecture and, as a result, we introduce the refined \textit{Segformer++} architecture. 
Our method can be applied not only at inference time without the need to re-train the model, but also to enhance the efficiency of the training process. \\[0.75ex]
The main contributions of this paper are:
\begin{itemize}[leftmargin=0.5cm,itemsep=0pt, topsep=-0.5\parskip]
    \item We show how to adapt the token merging strategy to transformers specialized in dense pixel tasks like Segformer.
    \item We present the \textit{Segformer++} architecture that makes optimal use of token merging in each stage of the Segformer.
    \item We evaluate the token merging strategy on multiple semantic segmentation and human pose estimation tasks and discuss the performance-speedup tradeoffs. 
\end{itemize}

\section{Related Work}
In dense pixel tasks like semantic segmentation and human pose estimation, an essential architectural requirement is the ability to work with high resolution images to ensure that small details in the images can be detected, and precise predictions can be made.
This usually comes with a high computational cost. 
Thus, a suitable trade-off between computational cost and performance has to be found.
A common strategy used in dense pixel tasks is to make use of a hierarchical pyramid structure by combining the feature maps of different stages for the final predictions \cite{pspnet, upernet, deeplabv3plus}.
In contrast, \cite{hrnet,hrnet_segmentation,hrnet_hpe} use multiple different branches with different feature map resolutions in parallel.
Although high quality results can be obtained with this method, the computational complexity is higher compared to pyramid networks. 
Therefore, transformer based approaches usually rely on the pyramid structure. \\[0.75ex]
In \cite{transformer} the transformer architecture is introduced and \cite{visiontransformer} first applies it to image processing.
While the attention mechanism ensures a global receptive field, the quadratic complexity of the self-attention mechanism makes it hard to apply the plain transformer architecture to high resolution images. 
Therefore, many architectures have been proposed that reduce the computational complexity of the attention mechanism. 
In \cite{swintransformer,multiscalevit,hrformer} the authors propose a hierarchical transformer architecture and reduce the computational complexity by using a shifted window attention mechanism. 
However, a major drawback of these methods is the loss of the global receptive field.
In \cite{pvt} a hierarchical structure is introduced together with an efficient attention mechanism that still features a global receptive field while significantly reducing the computational cost and in \cite{segformer} these ideas are incorporated into the Segformer architecture.
Nevertheless, further innovations are needed to make computations on high resolution images feasible.  
Therefore, in this paper we explore a strategy to speed up the Segformer architecture by merging similar tokens, thus making a step towards applying transformer based architectures in real-time applications and on edge devices. \\[0.75ex]
Since the quadratic cost of the attention mechanism is a major drawback of the transformer architecture, a lot of research, mostly in the context of image classification, has been done to reduce the effective number of tokens.
In \cite{tokenpruning1,tokenpruning2,tokenpruning3} token pruning strategies are explored. 
In token pruning, unimportant and redundant tokens in the input sequence are detected and simply removed, which, however, leads to losing their information.
To prevent this information loss, \cite{tokenpooling,tokencombining1,tokencombining2} combine similar tokens instead. 
In \cite{tokenmerging} a token merging strategy is introduced. 
The authors present an efficient way to calculate similarity scores between tokens and merge them based on these scores. 
This way, token merging can be used with any trained transformer without the need to re-train the model. 
Consequently, the efficiency of the ViT architecture is significantly increased for image classification tasks.
However, it is not straightforward how to apply this strategy to dense pixel tasks. 
Especially the hierarchical structure and the use of convolutional layers in the Segformer architecture make it hard to apply the token merging strategy directly. 
We adjust the token merging algorithm to the Segformer architecture and use a similar approach as in \cite{tokenmerging_sd}, where the algorithm is adapted to the Stable Diffusion architecture. \\[0.75ex]
Most similar to our approach is the work of \cite{pruningsegformer}, where a token pruning strategy for the Segformer architecture is implemented. 
Unlike our \textit{Segformer++}, they introduce additional learnable parameters, necessitating retraining of their model.
We will show the superior performance of our approach in the experiments section.

\section{Methods}
\subsection{Segformer Architecture}
\label{sec:segformer}
The Segformer architecture, introduced in \cite{segformer}, consists of a transformer encoder called MixTransformer (MiT) and a lightweight convolutional decoder. 
In contrast to the standard ViT, the MiT introduces some core modifications to increase the efficiency for dense prediction tasks. 
In this section we discuss these modifications. \\[0.75ex]
While the standard ViT extracts non-overlapping patches of size $16 \times 16$ pixels, the MiT extracts smaller overlapping patches of size $7 \times 7$ pixels. 
This way, more fine grained information in the images can be captured, which is crucial for dense prediction tasks.
However, this modification also increases the number of tokens, thus, raising the computational burden. 
Therefore, two additional modifications are introduced to reduce the computational cost of the attention: The pyramid structure and a efficient attention mechanism. \\[0.75ex]
The MiT model employs a pyramid structure to compute multiscale features across four stages 
Each stage generates a feature map of dimensions $\frac{H}{2^{i+1}} \times \frac{W}{2^{i+1}} \times D_i$ with $i \in \{1,2,3,4\}$. $H$, $W$, and $D_i$ are the image's height, image's width and the number of channels after each stage $i$.
To obtain the pyramid structure, the feature map resolution is reduced by a $3 \times 3$ convolution with stride $2$ in each stage.
Ultimately, the 4 feature maps undergo further processing by the decoder to yield the final predictions. \\[0.75ex] 
In the standard self-attention mechanism, the dimension of query $Q$, key $K$ and value $V$ is $N \times D$, where $N$ is the number of tokens and D is the embedding dimension of the tokens. 
To reduce the number of tokens, Segformer implements the Spatial Reduction Attention from \cite{pvt}.
This reduction is realized by applying a 2D convolution with stride $R$ to the keys and values before the attention mechanism.
We note that before the application of this convolution, $K$ and $V$ of shape $N_i \times D_i$ (with $N_i = \frac{H}{2^{i+1}} \cdot \frac{W}{2^{i+1}}$) are reshaped to $\frac{H}{2^{i+1}} \times \frac{W}{2^{i+1}} \times D_i$ and after the convolution the tokens are flattened again.
Consequently, the shape of $K$ and $V$ is reduced to $\frac{N_i}{R^2} \times D_i$ with the stride $R$ being called the reduction factor. \\[0.75ex]
In traditional ViTs, a MLP block consisting of 2 fully connected layers (FFNs) is used after the attention block to process the tokens in the channel dimension.
In contrast to that, the MiT introduces an additional $3 \times 3$ convolutional layer to include spatial information as $\text{MLP} = \text{FFN} \left( \text{Conv} \left( \text{FFN}(x) \right) \right)$. \\[0.75ex]
To conclude, the modifications increase the efficiency of the Segformer by reducing the number of tokens. 
The implementation of these modifications relies heavily on convolutional layers and the 2D structure of the tokens, thus, preventing the application of vanilla token merging.

\subsection{Token Merging Strategies}
\begin{figure*}
    \vspace{-0.3cm}
    \centering
    \begin{subfigure}{0.20\textwidth}
        \centering
        \includegraphics[width=\textwidth]{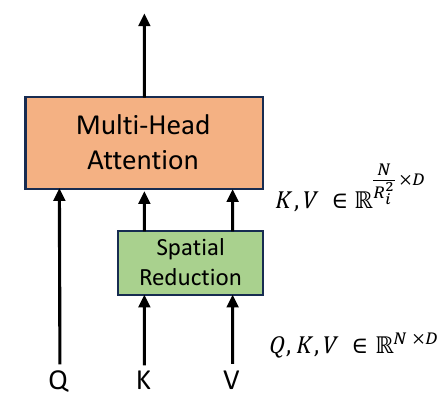}
        \caption{Spatial Reduction Attention used in Segformer \cite{segformer}}
        \label{fig:spatialreduction_attention}
    \end{subfigure}
    \hfill
    \begin{subfigure}{0.20\textwidth}
        \centering
        \includegraphics[width=\textwidth]{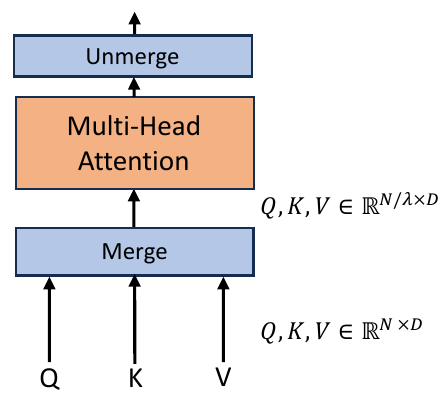}
        \caption{Token merging for Stable Diffusion \cite{tokenmerging_sd}}
        \label{fig:tome_attention}
    \end{subfigure}
    \hfill
    \begin{subfigure}{0.20\textwidth}
        \centering
        \includegraphics[width=\textwidth]{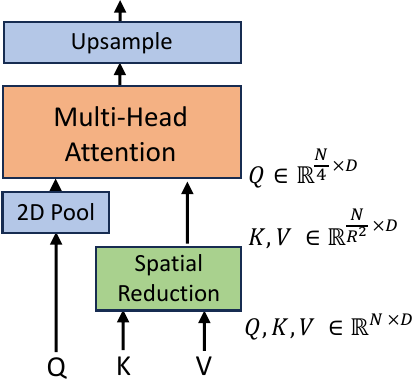}
        \caption{2D Neighbor Merging strategy (ours)}
        \label{fig:neighbor_attention}
    \end{subfigure}
    \hfill
    \begin{subfigure}{0.20\textwidth}
        \centering
        \includegraphics[width=\textwidth]{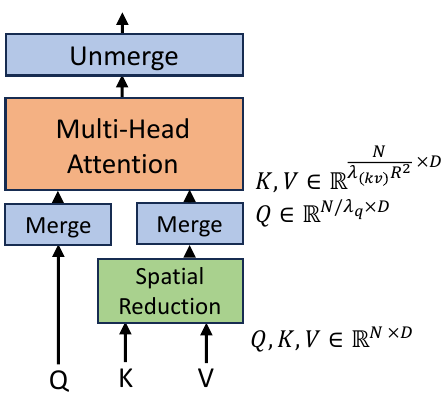}
        \caption{\textit{Segformer++} Attention (ours)}
        \label{fig:our_attention}
    \end{subfigure}
    \caption{Visualization of efficient attention mechanisms}
    \label{fig:attention}
    \vspace{-0.48cm}
\end{figure*}
In \cite{tokenmerging} a token merging strategy is introduced. 
It decreases the computational complexity of the attention mechanism by gradually reducing the number of tokens at runtime. 
In contrast to previous methods, the tokens are not pruned, but similar tokens are merged instead. 
This way, great performance on the ImageNet classification tasks can be achieved with huge efficiency gains. 
An important advantage of token merging is that existing models do not have to be re-trained. 
In this section, we describe the core ideas of token merging and how we adapt it for dense pixel tasks. \\[0.75ex]
The core idea of the original token merging paper \cite{tokenmerging} is to merge a fixed number $\tilde{r}$ of the most similar tokens in each layer, thus, iteratively reducing the number of token further with each layer. 
The parameter $\tilde{r}$ is called the reduction quantity and a larger $\tilde{r}$ results in a higher speedup but also larger performance losses.
To merge similar tokens, Bipartite Soft Matching is used. 
Therefore, the tokens are split into two groups $A$ and $B$ and the similarity score between each token in group $A$ and each token in group $B$ is calculated as
\begin{equation}
    \label{eq:similarity_score}
    \text{similarity}(A,B) = A \cdot B^T .
\end{equation}
Using these similarity scores, the $\tilde{r}$ most similar tokens are merged from group $B$ into group $A$. \\[0.75ex]
While this strategy works well for image classification tasks, it is not directly applicable to dense pixel tasks. 
Because the number of tokens is iteratively reduced in each layer, the 2D structure of the tokens is lost. 
However, this structure is pivotal in dense pixel tasks and building blocks like pyramid structures and convolutional layers rely on it.
In \cite{tokenmerging_sd} the token merging strategy is adapted to the Stable Diffusion architecture. 
They preserve the 2D structure by adjusting the algorithm such that the tokens are merged before the attention computation and directly unmerged afterwards. 
However, since this method does not iteratively increase the number of merged tokens, a more aggressive merging approach is necessary. 
Thus, instead of the reduction quantity $\tilde{r}$, a reduction rate $r$ is used. 
This rate denotes the percentage of merged tokens, e.g. $r=0.5$ corresponds to merging $50\%$ of the token, reducing the token number by the factor $\lambda = \frac{1}{1-r}=2$. \\[0.75ex]
Similar to \cite{tokenmerging_sd}, we repeatedly merge and unmerge the tokens in \textit{Segformer++}. 
Merging is computed via averaging and unmerging is done by copying a token to all positions that were merged into it.
Since the number of tokens in dense pixel tasks is usually significantly higher than in \cite{tokenmerging_sd}, we combine token merging with Segformer's Spatial Reduction Attention. 
This strategy is illustrated in Figure \ref{fig:our_attention}. \\[0.75ex]
We regard token merging as a smart merging strategy, since it is able to detect similar tokens. 
This way it is likely that small objects do not vanish as a result of the merging process. 
In contrast to that, we additionally introduce a simple merging strategy denoted as \textit{2D Neighbor Merging} that only merges neighboring tokens.
Before the attention computation, we just perform the Spatial Reduction for keys and values, while the number of queries is reduced by a 2D average pooling layer with stride 2 and pooling kernel 2, resulting in a $75\%$ reduction in the number of queries. 
This is illustrated in Figure \ref{fig:neighbor_attention}.
Comparing \textit{Segformer++} to \textit{2D Neighbor Merging} enables us to understand the effects of different merging strategies.  

\vspace{-0.01cm}
\subsection{Segformer++}
\begin{figure}[t]
    \vspace{-0.2cm}
    \begin{minipage}[b]{0.44\linewidth}
        \centering
        \includegraphics[width=\linewidth]{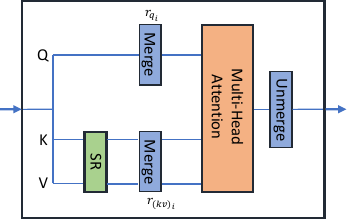}
        \captionof{figure}{Different reduction rates $r_{q_i}$ and $r_{(kv)_i}$}
        \label{fig:segformer++}
    \end{minipage}
    \hfill
    \begin{minipage}[b]{0.52\linewidth}
        \centering
            \resizebox{\linewidth}{!}{
                \begin{tabular}{|c|c|c|c|c|}
                    \hline
                    & \multicolumn{2}{c|}{Segformer++\textsubscript{HQ}} & \multicolumn{2}{c|}{Segformer++\textsubscript{fast}} \\
                    \hline
                    \textbf{Stage} $i$ & $r_{q_i}$ & $r_{(kv)_i}$ & $r_{q_i}$ & $r_{(kv)_i}$ \\
                    \hline
                    1 & 0 & 0.6 & 0 & 0.9\\
                    2 & 0 & 0.6 & 0 & 0.9\\
                    3 & 0.8 & 0 & 0.9 & 0 \\
                    4 & 0.8 & 0 & 0.9 & 0 \\
                    \hline
                \end{tabular}
            }
        \captionof{table}{Optimal reduction rates}
        \label{tab:rates}
    \end{minipage}
\vspace{-0.55cm}
\end{figure}

In \textit{Segformer++} we combine token merging with the hierarchical pyramid structure and the efficient attention mechanism of the Segformer architecture. 
Our architecture is identical to the original Segformer architecture, except for the attention where we apply token merging after the Spatial Reduction mechanism and unmerge the tokens after the attention. This is illustrated in Figure \ref{fig:our_attention}. 
Hence, the original weights published in \cite{segformer} can be used, but a significant speedup is obtained. \\[0.75ex]
Due to the hierarchical nature of the Segformer architecture, the computational cost varies significantly between the different stages.
Therefore, instead of always applying the same reduction rate $r$, we define a different rate in each stage $i$. 
Moreover, we use a different reduction rate in the merging of the queries than in the merging of the keys and values.
This is illustrated in Figure \ref{fig:segformer++} where the reduction rate for the queries is denoted as $r_{q_i}$ and for the keys and values as $r_{(kv)_i}$. \\[0.75ex]
We conduct multiple experiments to find the optimal rates for each stage. 
By using different reduction rates per stage and treating the queries differently than the keys and values, significantly better performance-speedup tradeoffs are achieved.
We note that the optimal values for $r_{q_i}$ and $r_{(kv)_i}$ depend on the image resolution.
As the majority of our experiments utilize images from the Cityscapes dataset with a resolution of $1024 \times 1024$ pixels, we optimize the rates accordingly.
As a result, we define two model variants with different performance-speedup tradeoffs: The \textit{Segformer++\textsubscript{HQ}} model achieves a good speedup with nearly no performance loss, while the \textit{Segformer++\textsubscript{fast}} model achieves a huge speedup with small performance losses.
The reduction rates for these models are given in Table \ref{tab:rates}.

\subsection{Computational Complexities}
In this section, we compare the computational complexities of the different attention mechanisms discussed in this paper. \\
Given queries $Q$, keys $K$ and values $V$ of shape $N \times D$, the vanilla attention mechanism can be computed as 
\begin{equation}
    {\small
    \text{attn(Q, K, V)} = \text{softmax}\left( \frac{QK^\mathrm{T}}{\sqrt{D}} \right) V
    }
\end{equation}
and, consequently, has a computational complexity of 
\begin{equation}
    O(N^2D) .
\end{equation}
In dense pixel tasks, the number of tokens $N$ is usually very high, thus, the main goal of efficient attention mechanisms is to reduce this number. \\[0.75ex]
In Figure \ref{fig:spatialreduction_attention} the Spatial Reduction Attention used in the Segformer architecture \cite{segformer} is depicted. Before the attention computation, the number of keys and values is reduced by a 2D convolution with stride $R$ such that their new shape is $\frac{N}{R^2} \times D $. 
Since the shape of the queries is left unchanged, the computational complexity of the attention computation is
\begin{equation}
    {\small
    O \left( \frac{N^2}{R^2}D \right) 
    }.
\end{equation}
As a result, the computational burden is reduced by the factor $R^2$.
We note that the computation of the strided convolution is negligible as it scales linearly in the number of tokens. \\[0.75ex]
In \cite{tokenmerging_sd}, token merging is adapted to the Stable Diffusion architecture as shown in Figure \ref{fig:tome_attention}. 
The token number is reduced by the factor \\[-1.75ex] 
\begin{equation}
    {\small
    \lambda = \frac{1}{1-r}
    }
\end{equation}
before the attention, resulting in the shape $\frac{N}{\lambda}\times D$ for queries, keys and values. Consequently, the computational complexity of the attention computation is $O(\frac{N^2}{\lambda^2}D)$. 
However, the token merging algorithm comes not for free due to the similarity score calculation in Equation \ref{eq:similarity_score}. If we assume that group $A$ and $B$ are of equal size, the computation scales as $O(\frac{N^2}{4}D)$. 
It is worth noting that, in practice, the group sizes are not equal, resulting in an even lower complexity.
Adding this overhead, the computational complexity is
\begin{equation}
    {\small
    O(\frac{N^2}{\lambda^2}D) + O(\frac{N^2}{4}D) = O((\lambda^{-2}+0.25)N^2D) 
    }
\end{equation}
and the computational expenses are reduced by the factor $\left( \lambda^{-2}+0.25 \right)^{-1}$. \\[0.75ex]
In our proposed \textit{Segformer++} architecture, we combine the token merging strategy with the Spatial Reduction Attention as illustrated in Figure \ref{fig:our_attention}. 
First, the number of keys and values is reduced by a strided convolution to the shape $\frac{N}{R^2} \times D$ while the queries are left unchanged. 
Next, token merging is applied to further reduce the number of tokens, resulting in the shape $\frac{N}{\lambda_{(kv)} R^2} \times D$ for keys and values and $\frac{N}{\lambda_q} \times D$ for the queries. 
We note that it is necessary to apply the token merging twice, since the queries cannot be treated similarly to the keys and values anymore. 
Thus, the complexity of the similarity score computation in Equation \ref{eq:similarity_score} is $O \left( \frac{N^2}{4}D \right) $ for the queries and $O \left( \frac{N^2}{4R^4}D \right)$ for the keys and values.
Based on the now reduced number of tokens, the computational complexity of the attention computation scales as $O(\frac{N^2}{\lambda_{(kv)} \lambda_q R^2}D)$ and the overall computational complexity can be expressed as
\begin{equation}
    {\small
    \begin{split}
    O \left( \frac{N^2}{\lambda_{(kv)} \lambda_q R^2}D \right) + O \left( \frac{N^2}{4}D \right) + O \left( \frac{N^2}{4R^4}D \right) \\ 
    = O \left( \left( \frac{1}{\lambda_{(kv)} \lambda_qR^2} + 0.25\frac{1+R^4}{R^4}  \right) N^2D \right) .
    \end{split}
    }
\end{equation}
Thus, the computational cost is reduced by the factor {$\left( \frac{1}{\lambda_{(kv)} \lambda_q R^2} + 0.25\frac{1+R^4}{R^4} \right)^{-1}$}. \\[0.75ex]
In conclusion, we see that \textit{Segformer++} is able to achieve superior speedups in theory.
We show in our experiments, that this strategy not only significantly reduces the computational burden, but also maintains the performance very well.

\section{Experiments}
We conduct multiple experiments to evaluate the performance of the proposed \textit{Segformer++} architecture. 
Our main focus is on semantic segmentation, but we also include some human pose estimation experiments to prove that our architecture can be used as a general purpose method for dense pixel tasks. 

\subsection{Experimental setup}
The semantic segmentation experiments are conducted on the Cityscapes dataset \cite{cityscapes} and the ADE20K dataset \cite{ade20k}, while the human pose estimation results are calculated on the MS COCO dataset \cite{coco} and the Jumping Broadcast Dataset (JBD) \cite{jumpdataset}. 
Special focus is given to Cityscapes and JBD due to their relatively high image resolutions, allowing for insights relevant to high-resolution scenarios.
For Cityscapes we use crops of size $1024 \times 1024$ pixels and the JBD is evaluated using crops of size $640 \times 480$ pixels. 
Although the images in ADE20K and MS COCO are not high resolution, we include the results to demonstrate the effectiveness of our method across diverse datasets. We note that efficiency measures on these datasets are not very insightful due to the low resolution. \\[0.75ex]
In addition to standard performance metrics, we particularly focus on model efficiency, defining the \textit{speedup} as
\begin{equation}
    {\small
    \text{speedup} = \frac{t_\text{orig}}{t_\text{mod}}
    } .
\end{equation}
Thus, it is the rate of the inference time of the original model $t_\text{orig}$ to the inference time of the modified model $t_\text{mod}$.
We measure the inference time using random tensors of a specific resolution. 
This way, we are also able to report speedup values for very high resolution data, even though no such datasets are publicly available.
Furthermore, we also report the speed and memory resources during training on real data.
To ensure a fair comparison, we train and evaluate each model on a single A100 GPU.
As main performance metrics, we use the Percentage of Correct Keypoints (PCK) for human pose estimation and the mean Intersection over Union (mIoU) for semantic segmentation.
Additionally, we define the mIoU\textsubscript{small} metric on the Cityscapes dataset to specifically assess segmentation performance on small object classes.
Thus, we see whether merging has a detrimental impact on small objects. 
The classes used for the mIoU\textsubscript{small} computation are \textit{Fence, Pole, Traffic Light, Traffic Sign}. \\[0.75ex]
Two variants of the \textit{Segformer++} architecture are tested: The \textit{Segformer++\textsubscript{HQ}} model that enables a good speedup with nearly no performance loss, and the \textit{Segformer++\textsubscript{fast}} model that enables a huge speedup at a small performance loss. 
Additionally, we always evaluate a model utilizing the \textit{2D Neighbor Merging} strategy for comparison.
Unless otherwise stated, all model variants are based on the Segformer-B5 architecture, which is the largest available Segformer model. \\[0.75ex]
All models are implemented in PyTorch \cite{pytorch}, and the MMsegmentation \cite{mmsegmentation} and MMpose \cite{mmpose} libraries are used for training and evaluation. Our code is publicly available at \url{https://kiedani.github.io/MIPR2024/}.

\subsection{Inference for Semantic Segmentation}
Our merging strategies offer the advantage of being applicable without requiring model re-training. Thus, in this section, we assess these strategies solely during inference. We utilize official weights from \cite{segformer} for the Segformer-B2 and our trained weights (see Section \ref{sec:segtrain}) for the Segformer-B5 model.
\begin{table}[h!t]
    \vspace{-0.1cm}
    \centering
    \resizebox{0.8\linewidth}{!}{
    \begin{tabular}{|c|c|c|c|c|}
        \hline
        & \multicolumn{4}{c|}{\textbf{Segformer-B5}} \\
        \hline
        Method                             & mIoU $\uparrow$  & mIoU\textsubscript{small} $\uparrow$ & FPS $\uparrow$   & Speedup $\uparrow$ \\
        \hline
        Segformer (original)               & 82.39 & 72.97                     & 14.75 & 1.00    \\
        Segformer++\textsubscript{HQ} (ours)             & 82.31 & 72.93                     & 23.68 & 1.61    \\
        Segformer++\textsubscript{fast} (ours)             & 82.04 & 72.72                     & 28.66 & 1.94    \\
        2D Neighbor Merging & 81.96 & 72.24                     & 28.08 & 1.90    \\
        Downsampling                       & 77.31 & 65.03                     & 96.08 & 6.51    \\
        \hline
        \hline
        & \multicolumn{4}{c|}{\textbf{Segformer-B2}} \\
        \hline
        Method                             & mIoU $\uparrow$  & mIoU\textsubscript{small} $\uparrow$ & FPS $\uparrow$ & Speedup $\uparrow$ \\
        \hline
        Segformer (original)               & 81.08 & 71.97                     & 36.92  &         \\
        Segformer++\textsubscript{HQ} (ours)             & 81.03 & 71.83                     & 56.29  & 1.52    \\
        Segformer++\textsubscript{fast} (ours)             & 80.62 & 71.53                     & 71.98  & 1.95    \\
        2D Neighbor Merging & 80.38 & 71.02                     & 67.86  & 1.84    \\
        Downsampling                       & 75.87 & 63.89                     & 243.15 & 6.59    \\
        Segformer (pruning) \cite{pruningsegformer}               & 80.03\textsuperscript{*} &     -     &    -    &    -   \\
        \hline
    \end{tabular}
    }
    \vspace{-0.1cm}
    \caption{Semantic segmentation inference on Cityscapes. The large Segformer-B5 models use batch size 8 and the smaller Segformer-B2 models use batch size 16. The * indicates that the results are taken from the original paper instead of our own experiments.}
    \label{tab:segcityscapes}
    \vspace{-0.25cm}
\end{table} \\
In the upper part of Table \ref{tab:segcityscapes} we show the results for the Cityscapes dataset. 
The \textit{Segformer++\textsubscript{HQ}} model is $61\%$ faster while maintaining a consistent mIoU performance.
Moreover, with only a slight performance loss, the Segformer++\textsubscript{fast} model is $94\%$ faster, and thus, is nearly twice as fast as the original Segformer model.
Even though the \textit{2D Neighbor Merging} achieves a similar mIoU performance as the \textit{Segformer++\textsubscript{fast}} model, we see differences in the mIoU\textsubscript{small} metric.
Consequently, we conclude that a smart token merging strategy is in fact able to better maintain the information of small objects.
We also provide some segmentation results in Figure \ref{fig:feature-graphic} to visualize the performance of our \textit{Segformer++} models.
The segmentation maps are nearly indistinguishable, even though the inference time is significantly reduced. 
Since the number of tokens can also simply be reduced by downsampling the image, we compare our models to this strategy, too.
We choose a resolution of $\frac{H}{2} \times \frac{W}{2}$ to achieve a similar token reduction as for the queries in \textit{2D Neighbor Merging}.
As expected, this strategy results in a significant performance loss, especially in the mIoU\textsubscript{small} metric. \\[0.75ex]
We also provide results computed with the smaller Segformer-B2 model in the lower part of Table \ref{tab:segcityscapes}.
These results are similar to the results of the larger Segformer-B5 model.
We additionally include the results of \cite{pruningsegformer} in the table to compare our merging strategies to their token pruning approach.
However, it becomes pretty clear that our method is superior to the token pruning approach, especially since their approach requires an additional re-training of the model. 

\subsection{Training for Semantic Segmentation}
\label{sec:segtrain}
Additionally, to inference time only usage, it is also possible to apply our strategies during training and, thus, to reduce the computational cost and the memory resources.
Thus, larger models can be trained on less expensive hardware and in shorter time, which is especially interesting for small research labs and practitioners.  
\begin{table}[ht]
    \vspace{-0.1cm}
    \centering
    \resizebox{0.8\linewidth}{!}{
        \begin{tabular}{|c|c|c|c|c|}
            \hline 
             & \multicolumn{4}{c|}{\textbf{Cityscapes}} \\
            \hline
            Method                                 & mIoU $\uparrow$  & mIoU\textsubscript{small} $\uparrow$ & Steps/s $\uparrow$  & Memory (GB) $\downarrow$ \\
            \hline
            Segformer (original)                   & 82.39 & 72.97                     & 0.80       & 48.30       \\
            Segformer++\textsubscript{HQ} (ours)   & 82.19 & 72.77                     & 1.12       & 33.95       \\
            Segformer++\textsubscript{fast} (ours) & 81.77 & 72.39                     & 1.24       & 30.50       \\
            2D Neighbor Merging                    & 82.38 & 72.81                     & 1.30       & 31.10       \\
            Downsampling                           & 79.24 & 67.75                     & 2.36       & 10.00       \\
            \hline
            \hline
             & \multicolumn{4}{c|}{\textbf{ADE20K}} \\
            \hline
            Method                                 & mIoU $\uparrow$  & mIoU\textsubscript{small} $\uparrow$ & Steps/s $\uparrow$ & Memory (GB) $\downarrow$ \\
            \hline
            Segformer (original)                   & 49.72 & -                         & 1.17       & 33.68       \\
            Segformer++\textsubscript{HQ} (ours)   & 49.77 & -                         & 1.34       & 29.18       \\
            Segformer++\textsubscript{fast} (ours) & 49.10 & -                         & 1.40       & 28.04       \\
            2D Neighbor Merging                    & 49.35 & -                         & 1.47       & 27.17       \\
            Downsampling                           & 46.71 & -                         & 2.21       & 12.41       \\
            \hline
        \end{tabular}
    }
    \vspace{-0.1cm}
    \caption{Semantic segmentation training on Cityscapes and ADE20K. Computed with Segformer-B5 using a batch size of 4 for Cityscapes and 8 for ADE20K.}
    \label{tab:seg-train}
    \vspace{-0.25cm}
\end{table} \\
In Table \ref{tab:seg-train} we show the results on the Cityscapes dataset. 
The Segformer++ models achieve good speedups and a significant memory reduction, while maintaining very good performance on the mIoU as well as the mIoU\textsubscript{small} metric.
Moreover, \textit{2D Neighbor Merging} slightly outperforms all other models.
Thus, it seems like the \textit{2D Neighbor Merging} model is able to learn how to deal with small objects during training, even though it could not handle these as well in the inference only setting.
Furthermore, by evaluating the results on the ADE20K dataset in the lower part of Table \ref{tab:seg-train}, we find that Segformer++ slightly outperforms the \textit{2D Neighbor Merging} strategy.
Consequently, both strategies proposed in this paper are able to achieve very good performance values during training. Since the \textit{Segformer++\textsubscript{HQ}} model even slightly outperforms the original Segformer model on the ADE20K dataset, we think that our method can be always used as a drop-in replacement for the original Segformer model. \\[0.75ex]
In conclusion, our strategies all yield very good performance values and the differences between the models are minimal.
As a result, we provide fast and high performing models that can be easily finetuned on private datasets using less expensive hardware, making it a very valuable tool for researchers and practitioners interested in efficiency.

\subsection{Training on Human Pose Estimation}
Even though the Segformer architecture is mainly designed for semantic segmentation tasks, it is straight forward to apply it to other dense pixel tasks. 
Therefore, we train our models for human pose estimation in this section. We perform our experiments using a Top-Down approach given ground truth bounding boxes. Our model is now adjusted such that it predicts heatmaps instead of segmentation maps. Consequently, we set the output dimension to the number of keypoints.
\begin{table}[ht]
    \vspace{-0.1cm}
    \centering
    \resizebox{0.8\linewidth}{!}{
    \begin{tabular}{|c|c|c|c|c|}
        \hline
        & \multicolumn{4}{c|}{\textbf{Jumping Broadcast Dataset}} \\
        \hline
        Method                                 & PCK@0.1 $\uparrow$ & PCK@0.05 $\uparrow$ & Steps/s $\uparrow$  & Memory (GB) $\downarrow$ \\
        \hline
        Segformer (original)                   & 95.20   & 90.65    & 1.10     & 40.00       \\
        Segformer++\textsubscript{HQ} (ours)   & 95.18   & 90.51    & 1.31     & 35.95       \\
        Segformer++\textsubscript{fast} (ours) & 94.58   & 89.87    & 1.37     & 34.58       \\
        2D Neighbor Merging                    & 95.17   & 90.16    & 1.40     & 33.37       \\
        \hline
        \hline
         & \multicolumn{4}{c|}{\textbf{MS COCO}} \\
        \hline
        Method                                 & PCK@0.1 $\uparrow$ & PCK@0.05 $\uparrow$ & Steps/s $\uparrow$   & Memory (GB) $\downarrow$ \\
        \hline
        Segformer (original)                   & 95.16   & 87.61    & 2.33      & 13.54       \\
        Segformer++\textsubscript{HQ} (ours)   & 94.97   & 87.35    & 2.27      & 13.12       \\
        Segformer++\textsubscript{fast} (ours) & 95.02   & 87.37    & 2.30      & 12.92       \\
        2D Neighbor Merging                    & 94.98   & 87.36    & 2.88      & 12.27       \\
        \hline
    \end{tabular}
    }
    \vspace{-0.05cm}
    \caption{Human Pose Estimation training. Results computed with Segformer-B5 and batch size 16.}
    \label{tab:hpe}
    \vspace{-0.25cm}
\end{table} \\
The results on the Jumping Broadcast Dataset in Table \ref{tab:hpe} show, that both the \textit{Segformer++\textsubscript{HQ}} and the \textit{2D Neighbor Merging} strategy achieve nearly the same performance as the original Segformer model. 
The \textit{Segformer++\textsubscript{HQ}} model, however, slightly outperforms the \textit{2D Neighbor Merging} on the PCK@0.05 metric due to the better handling of small objects.
All models achieve excellent results on JBD. 
\begin{wrapfigure}{r}{0.5\linewidth}
    \vspace{-0.2cm}
    \centering
    \includegraphics[width=\linewidth]{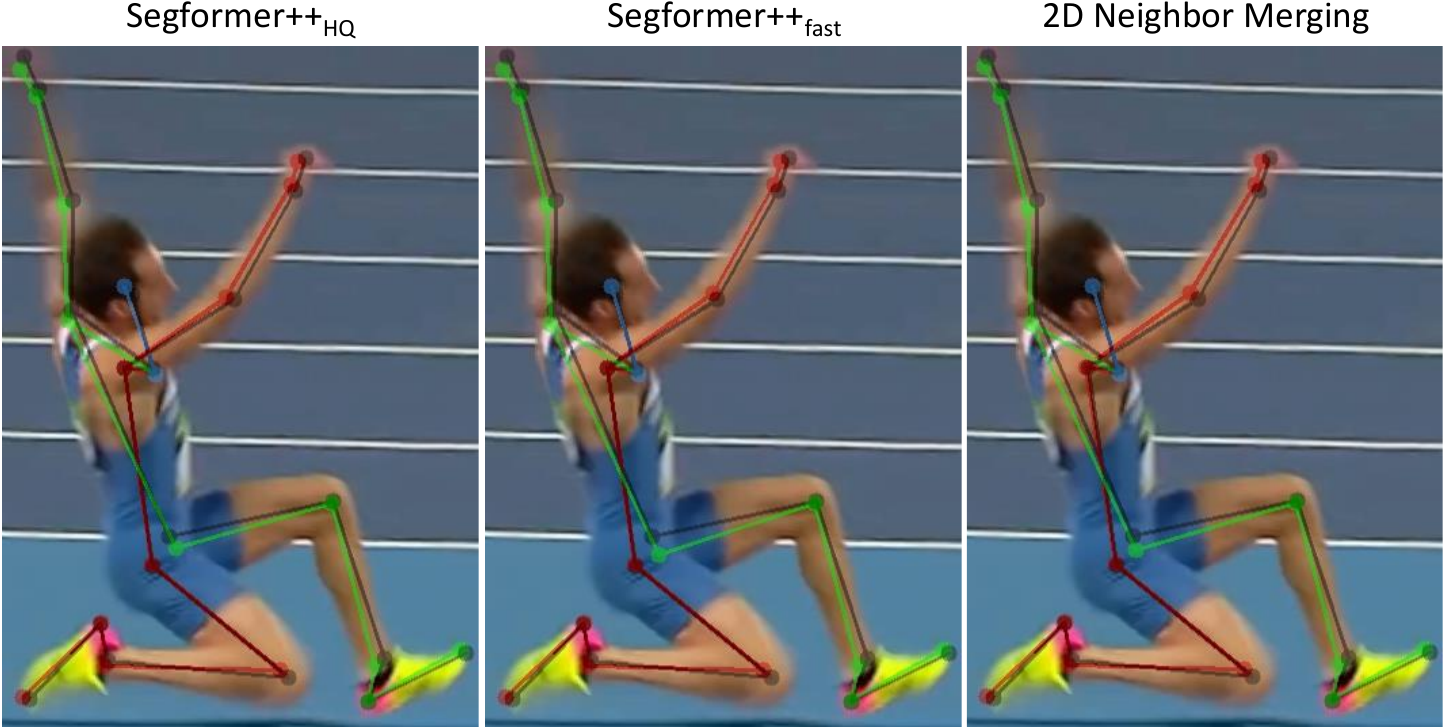}
    \caption{Example predictions on JBD. The ground truth is visualized in black.}
    \label{fig:posen}
    \vspace{-0.25cm}
\end{wrapfigure}
Moreover, the models also yield great results on the MS COCO dataset, thus, validating the results of the segmentation experiments for human pose estimation tasks. 
We illustrate some example predictions in Figure \ref{fig:posen} to visualize the performance of our models.
Similar to the semantic segmentation visualization, the predictions are nearly indistinguishable, even though the inference time is significantly reduced. \\[0.75ex]   
In conclusion, all strategies achieve very good performance values with the \textit{Segformer++\textsubscript{HQ}} model slightly outperforming the other models on small details.
Thus, we prove that our models can easily be used as drop-in replacements for other dense pixel tasks as human pose estimation.

\subsection{Inference Speed for Different Resolutions}
While the resolution of Cityscapes and JBD is relatively high compared to other publicly available datasets, it is still way lower than the resolution images captured with modern smartphone cameras. 
Since there are no public datasets to evaluate the performance of our models on such high resolution images we cannot report corresponding performance metrics. 
However, we are confident that the trends observed in the previous experiments also hold for higher resolution images. 
Nevertheless, we can discuss the speedups achieved on higher resolution data by using random tensors of different resolutions as input to the models.
We note that neither the reduction rate nor the computational complexity of the matching process are dependent on the input content.
Thus, the speedups achieved on random tensors are representative for real data as well.
\begin{table}[ht]
    \vspace{-0.1cm}
    \centering
    \resizebox{0.9\linewidth}{!}{
    \begin{tabular}{|c|c|c|c|c|c|}
        \hline
        Method                                  & 512x512 & 640x640 & 1024x1024 & 2048x1024 & 3840x2160 \\
        \hline
        Segformer++\textsubscript{HQ} (ours)    & 1.18    & 1.32    & 1.61      & 2.04         & 2.66   \\
        Segformer++\textsubscript{fast} (ours)  & 1.27    & 1.45    & 1.94      & 2.75         & 4.31   \\
        2D Neighbor Merging                     & 1.38    & 1.50    & 1.90      & 2.39         & 3.23   \\
        \hline
    \end{tabular}
    }
    \vspace{-0.05cm}
    \caption{Speedups achieved using random generated tensor of various sizes. All speedups are relative to the original Segformer model. The results are computed with Segformer-B5 and the optimal batch size for each resolution is used.}
    \label{tab:speedups}
    \vspace{-0.25cm}
\end{table} \\
In Table \ref{tab:speedups} we show the speedup values achieved by our models on different input resolutions.
The \textit{2D Neighbor Merging} strategy is very efficient for lower resolutions, but it is outperformed by our Segformer++ models on higher resolution data. 
This is since the pooling in \textit{2D Neighbor Merging} is only applied to the queries as illustrated in \ref{fig:neighbor_attention}, while the token merging in our \textit{Segformer++} models is also applied to the keys and values. 
It is not possible to define the best model for all resolutions, but the \textit{Segformer++} models especially efficient on high resolution data.
Conclusively, \textit{Segformer++} is a valuable tool for speeding up the calculations on high resolution data.

\section{Conclusion and Outlook}
In this paper, we combine the Segformer architecture with token merging strategies.
We propose the \textit{Segformer++} architecture and the 2D Neighbor Merging strategy that are both able to significantly reduce the computational burden of the attention mechanism.
We show that significant speedups can be achieved while maintaining high performance values.
The models are especially well suited for high resolution data, where remarkable speedups can be achieved. \\[0.75ex]
The \textit{Segformer++} as well as the \textit{2D Neighbor Merging} strategy achieve very good performance values and it is not clear which strategy is superior.
However, as \textit{Segformer++} is able to achieve slightly better performance values on small objects in the inference only setting, we think that using this architecture is the best choice in most cases.
Consequently, it is a valuable tool for researchers and practitioners interested in efficient and high performance dense pixel tasks on high resolution data.
Our strategy enables training powerful models on less expensive hardware and makes running such models on embedded devices feasible. \\[0.75ex]
Even though we concentrate on the Segformer architecture in this paper, our strategies can easily be extended to other transformer architectures that both use convolutional layers and the attention mechanism.
We hope that our work enables other researchers to increase the efficiency of their models and, thus, make them more accessible to a broader audience.


\newpage
\bibliographystyle{IEEEtran}
\bibliography{main}

\end{document}